%% file: main.tex
\documentclass[10pt,twocolumn,letterpaper]{article}

\usepackage{cvpr}
\usepackage{times}
\usepackage{epsfig}
\usepackage{graphicx}
\usepackage{amsmath}
\usepackage{amssymb}
\usepackage{bm}
\usepackage{rotating}
\usepackage{mathtools}
\usepackage{hhline}
\newcommand{\R}{\mathbb{R}}

\usepackage[pagebackref=true,breaklinks=true,letterpaper=true,colorlinks,bookmarks=false]{hyperref}

\cvprfinalcopy 

\def\cvprPaperID{10682} 

\ifcvprfinal\fi
\begin{document}

\title{Interpretable and Accurate Fine-grained Recognition via Region Grouping}

\author{Zixuan Huang$^1$ \quad \quad  Yin Li$^{2,1}$\\
$^1$Department of Computer Sciences, $^2$Department of Biostatistics and Medical Informatics\\
University of Wisconsin--Madison\\
{\tt\small \{zhuang356, yin.li\}@wisc.edu}
}%

\maketitle
\thispagestyle{empty}

\begin{abstract}
\input{src/abstract.tex}
\end{abstract}

\input{src/intro.tex}

\input{src/related_work.tex}

\input{src/method.tex}

\input{src/exp.tex}

\input{src/conclusion.tex}

\bibliographystyle{ieee_fullname}
\bibliography{egbib_short}

\end{document}

%% file: src/abstract.tex
We present an interpretable deep model for fine-grained visual recognition. 
At the core of our method lies the integration of region-based part discovery and attribution within a deep neural network. 
Our model is trained using image-level object labels, and provides an interpretation of its results via the segmentation of object parts and the identification of their contributions towards classification. 
To facilitate the learning of object parts without direct supervision, we explore a simple prior of the occurrence of object parts. We demonstrate that this prior, when combined with our region-based part discovery and attribution, leads to an interpretable model that remains highly accurate. 
Our model is evaluated on major fine-grained recognition datasets, including CUB-200~\cite{wah2011caltech}, CelebA~\cite{liu2015deep} and iNaturalist~\cite{van2018inaturalist}. Our results compare favorably to state-of-the-art methods on classification tasks, and our method outperforms previous approaches on the localization of object parts. Our project website can be found at \url{https://www.biostat.wisc.edu/~yli/cvpr2020-interp/}.

%% file: src/intro.tex
\section{Introduction}
\begin{figure}
	\centering
	\includegraphics[width=0.95\linewidth]{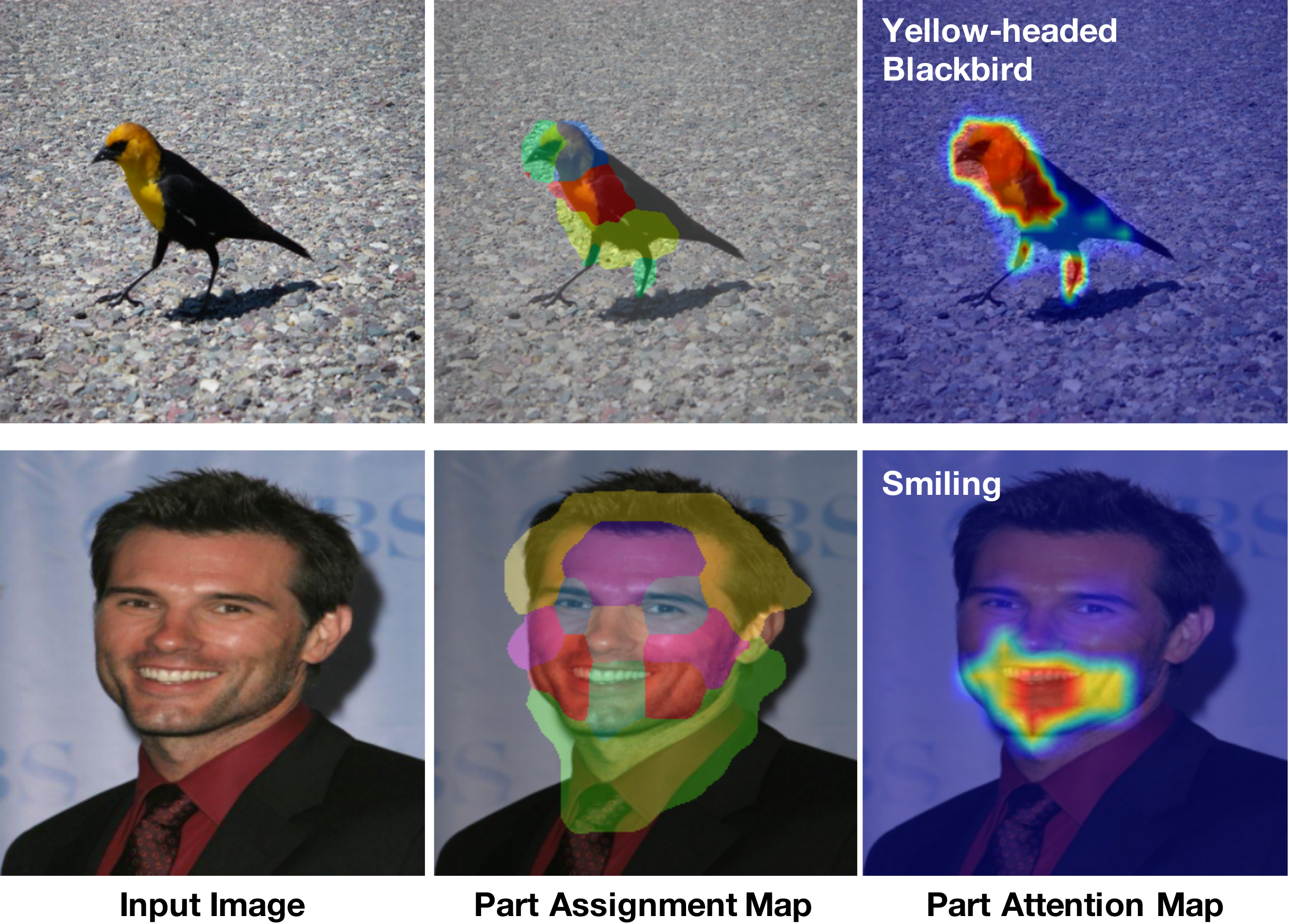}
    \caption{Why does a deep model recognize the bird as {\it ``Yellow-headed Blackbird''} or consider the person {\it ``Smiling''}? We present an interpretable deep model for fine-grained recognition. Given an input image (left), our model is able to segment object parts (middle) and identify their contributions (right) for the decision. {\it Results are from our model trained using only image-level labels}. }
    \label{fig:teaser}\vspace{-1.5em}
\end{figure}

Deep models are tremendously successful for visual recognition, yet their results are oftentimes hard to explain. Consider the examples in Fig.\ \ref{fig:teaser}. Why does a deep model recognize the bird as ``Yellow-headed Blackbird'' or consider the person ``Smiling''? While the interpretation of a model can happen at multiple facets, we believe that at least one way of explaining the model is to segment meaningful regions of object parts (e.g., the eyes, mouth, cheek, forehead and neck of a face), and further identify their contributions towards the decision (e.g., the mouth region is more discriminative for smiling). How can we design an interpretable deep model that learns to discover object parts and estimates their importance for visual recognition?

It turns out that part discovery, i.e., learning object parts without explicit supervision of part annotations, is by itself a challenging problem. As a baby step, we focus on the task of fine-grained recognition, where the parts belonging to the same super category share common visual patterns. For example, most tails of birds have a similar shape. Our key observation is that features from a convolutional network can be used to group pixels into a set of visually coherent regions~\cite{jampani2018superpixel,hung2019scops}, from which a subset of discriminative segments can be selected for recognition~\cite{li2018beyond,li2019expectation,chen2019graph}. \emph{With only object labels as the guidance}, we hope that the grouping will help to find visually distinct parts, and the selection process will identify their contributions for classification. 

A major challenge for our region-based part discovery is that there is no explicit supervisory signal to define part regions. Therefore, prior knowledge about object parts must be incorporated to facilitate the learning. A core innovation of our work is the exploration of a simple prior about object parts: given a single image, the occurrence of a part follows a U-shaped distribution. For example, the head of a bird is likely to occur in most bird images while the legs of a bird might only appear in some images. Surprisingly, we demonstrate that this simple prior, when combined with our region-based part discovery, leads to the identification of meaningful object parts. More importantly, the resulting interpretable deep model remains highly accurate. Several recent methods have been developed for discovering parts in fine-grained classification, yet none of them considered the prior we use. 

To this end, we present our interpretable deep model for fine-grained classification. Specifically, our model learns a dictionary of object parts, based on which a 2D feature map can be grouped into ``part'' segments. This is done by comparing pixel features to part representations in a learned dictionary. Moreover, region-based features are pooled from the result segments, followed by an attention mechanism to select a subset of segments for classification. Importantly, during training, we enforce a U-shaped prior distribution for the occurrence of each part. This is done by minimizing the Earth Mover's Distance between our prior and the empirical distribution of part occurrence. During training, our model is only supervised by object labels with our proposed regularization term. During testing, our model jointly outputs the segments of object parts, the importance of the segmented parts, and the predicted label. The interpretation of our model is thus granted by the part segmentation and the contribution of each part for classification. 

To evaluate our model, we conduct extensive experiments using three fine-grained recognition datasets for both interpretability and accuracy. To quantify interpretability, we compare the output region segments from our model to the annotated object parts. For accuracy, we report standard metrics for fine-grained classification. On smaller scale datasets, such as CUB-200~\cite{liu2015deep} and CelebA~\cite{wah2011caltech}, our model is shown to find parts of the birds and faces with low localization error, while at the same time compares favorably to state-of-the-art methods in terms of accuracy. On the more challenging iNaturalist dataset~\cite{van2018inaturalist}, our model improves the accuracy of a strong baseline network (ResNet101) by \textbf{5.7\%}, reduces the object localization error, and demonstrates promising qualitative results for part discovery.

%% file: src/related_work.tex
\section{Related Work}
There has been an emerging interest in explaining deep models. Our work focused on developing interpretable deep models for fine-grained classification, following the paradigm of region-based recognition. We briefly survey relevant literature on interpretable deep learning, part-based fine-grained classification, and recent deep models for region segmentation and region-based recognition.

\noindent \textbf{Visualizing and Understanding Deep Networks}. 
Several recent efforts have been developed to visualize and understand a trained deep network. Many of these post-hoc approaches~\cite{mahendran2015understanding, erhan2009visualizing, zeiler2014visualizing, dosovitskiy2016inverting, zhou2014object, simonyan2013deep} focused on developing visualization tools for the activation maps and/or the filter weights within trained networks. Other works sought to identify the discriminative regions in an input image given a pre-trained network~\cite{sundararajan2017axiomatic, selvaraju2017grad, simonyan2013deep, fong2017interpretable, zeiler2014visualizing, petsiuk2018rise, bau2017network, zhou2016learning,lundberg2017unified}. Beyond qualitative results, Bau et al.\ \cite{bau2017network} proposed a quantitative benchmark that compares the activation of a network's units to human-annotated concept masks. An alternative direction is to learn a simple model, such as a linear classifier~\cite{ribeiro2016should} or decision tree~\cite{craven1996extracting}, to mimic the behavior of a trained network, thereby providing an explanation of the target model's outputs. Our work shares the same motivation for interpreting deep models, yet we integrate interpretation into the learning of the model. Similar to~\cite{bau2017network}, we also use human-annotated object parts to quantify the interpretability of our network.

\noindent \textbf{Interpretable Deep Models}.
Interpretability can be built with a deep model. Many recent works developed deep models that are interpretable by their design. For instance, Zhang et al.\ \cite{zhang2018interpretable} designed a regularization method that encourages each filter in high-level convolutional layers to focus on a specific object part. Brendel et al.\ \cite{brendel2019approximating} proposed BagNet that takes small image patches as input, followed by a bag-of-feature (BoF) representation for whole image classification. BagNet can naturally attribute the decision to local regions, and thus help to explain the decision process. Alvarez-Melis and Jaakkola~\cite{melis2018towards} proposed to assign relevance scores to the basis of global image features. Alternatively, new network architectures can be designed for interpretable models. For example, Capsule Networks~\cite{sabour2017dynamic} substitute the commonly used scalar activation with vectors, where the latter is believed to represent entities such as an object or an object part. A relevant idea is further extended in~\cite{Sun_2019_ICCV} upon conventional CNNs by enforcing sparse connection from convolutional units to the final prediction. 

The most relevant work is from Chen et al.\ \cite{chen2018looks}. They proposed to learn prototypes of object parts within the network. The decision of the model thus depends on the identification of the prototypes found in the input image. Similar to their work, our model also seeks to explicitly encode the concepts of object parts. However, our work is different from~\cite{chen2018looks} in two key aspects: (1) we adopt region grouping to provide explanation grounded to image segments; and (2) the learning of our model is regularized by a strong prior of the occurrence of object parts. 

\noindent \textbf{Part Discovery for Fine-grained Recognition}.
Identifying discriminative object parts is important for fine-grained classification~\cite{simon2014part, simon2015neural, xiao2015application, zhang2016picking}. For example, bounding box or landmark annotations can be used to learn object parts for fine-grained classification~\cite{huang2016part, lin2015deep, parkhi2011truth, zhang2016spda, zhang2014part}. To avoid costly annotation of object parts, several recent works focused on unsupervised or weakly-supervised part learning using deep models. Xiao et al.\ \cite{xiao2015application} performed spectral clustering on convolutional filters to find representative filters for parts. Wang et al.\ \cite{wang2018learning} proposed to learn a bank of convolutional filters that captures class-specific object parts. Moreover, attention models have also been explored extensively for learning parts. Liu et al.\ \cite{liu2016fully} made use of reinforcement learning to select region proposals for fine-grained classification. Zheng et al.\ \cite{zheng2017learning} grouped feature channels for finding parts and their attention, where channels sharing similar activation patterns were considered as a part candidate.

Similar to previous works, our work also seeks to find parts and to identify their importance for fine-grained classification. However, our work differs from previous works by considering an explicit regularization of the occurrence of object parts. Moreover, we also consider a large scale dataset (iNaturalist~\cite{van2018inaturalist}) for part discovery. We will compare to previous methods on both recognition accuracy and part localization error in our experiments.

\noindent \textbf{Weakly-supervised Segmentation of Object Parts}. Our work is also connected to previous works on weakly-supervised or unsupervised segmentation of object parts. Zhang et al.\ \cite{zhang2017growing} extracted activation from a pre-trained CNN to represent object parts in a graph. Their learning of parts is supervised by a few part annotations. Collins et al.~\cite{collins2018} performed a non-negative matrix factorization over the activation of a pre-trained CNN, where each component defines a segment of the image. Jampani et al.\ \cite{jampani2018superpixel} proposed an iterative deep model for learning superpixel segments. More recently, Hung et al.~\cite{hung2019scops} presented a deep model that incorporates strong priors, such as spatial coherence, rotation invariance, semantic consistency and saliency, for unsupervised learning of object parts. Our work is inspired by~\cite{hung2019scops}, where we also explore novel regularization for learning to segment object parts. However, we consider weakly supervised part segmentation in the context of fine-grained classification. Moreover, we explore a very different prior of part occurrence.

\noindent \textbf{Region-based Recognition}.
Finally, our model combines segmentation and classification into a deep model, thus links to the efforts of region-based recognition~\cite{gu2009recognition, yan2015object, kohli2009robust, achanta2012slic}, or compositional learning~\cite{stone2017teaching}. There has been a recent development of designing deep models for region-based recognition. For example, Li et al.\ \cite{li2018beyond} proposed to group CNN features into a region graph, followed by a graph convolutional network for visual recognition. Similar ideas were also explored by Chen et al.\ \cite{chen2019graph}. More recently, Li et al.\ \cite{li2019expectation} presented a deep model that jointly refines the grouping and the labeling of the regions, using expectation-maximization. Moreover, Arslan~\cite{arslan2018graph} proposed to build a graph neural network using pre-defined regions for brain image classification. Our model uses a similar idea to~\cite{li2018beyond, chen2019graph,li2019expectation} for grouping CNN features. However, none of these previous works focused on the quality of grouping, and thereby they can not be directly used for interpretation. 

%% file: src/method.tex
\section{Method}
\begin{figure*}[t]
\centering
	\includegraphics[width=0.9\linewidth]{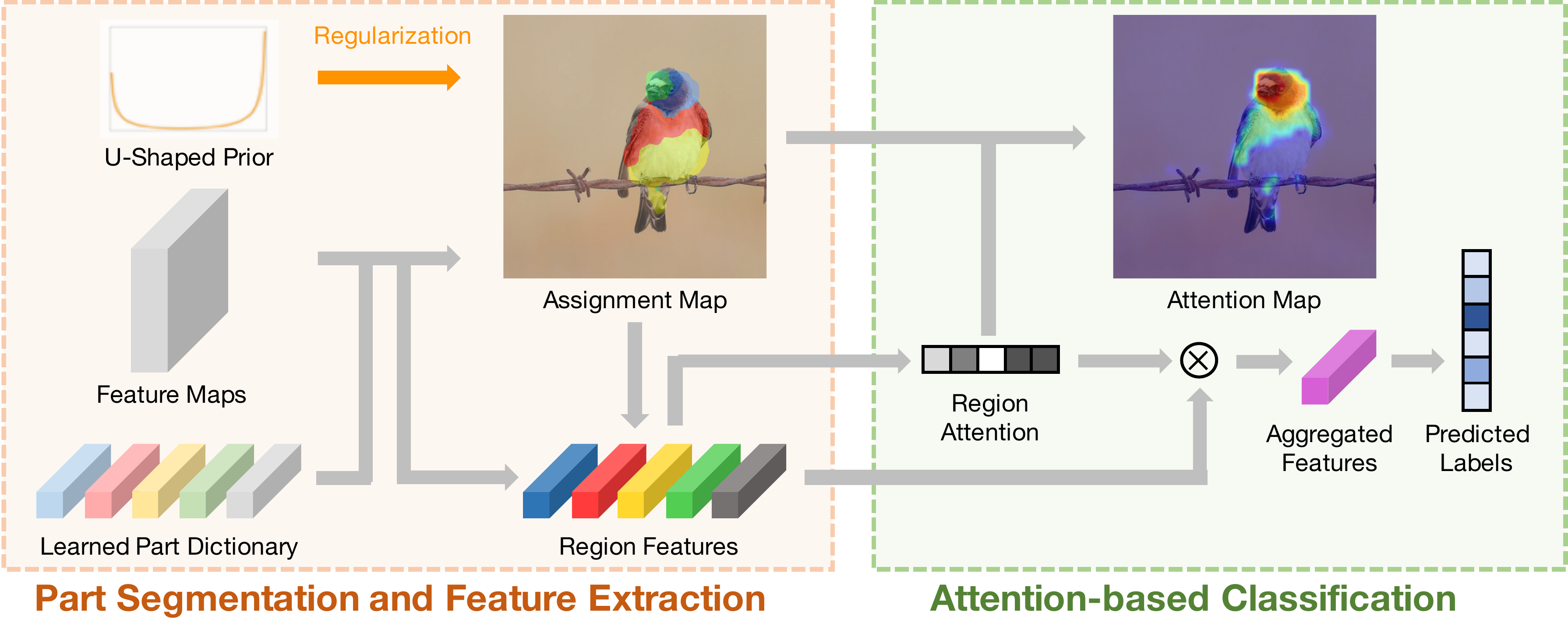}
	\caption{Overview of our method. With image-level labels, our model learns to group pixels into meaningful object part regions and to attend to these part regions for fine-grained classification. Our key innovation is a novel regularization of part occurrence that facilitates part discovery during learning. Once learned, our model can output (1) a part assignment map; (2) an attention map and (3) the predicted label of the image. We demonstrate that our model provides an accurate and interpretable deep model for fine-grained recognition.} \vspace{-1.5em}
	\label{fig:structure}
\end{figure*}

Consider a set of $N$ 2D image feature maps $\bm X_{1:N} = \{\bm X_n\}$ and their categorical labels $y_{1:N} = \{y_n\}$, where $\bm X_n \in \R^{D \times H \times W}$ is $D$-dimensional features on 2D image plane $H\times W$ from a convolutional network and $y_n \in [1, ..., c]$ is the image-level label of fine-grained categories. The goal of our model is to learn a part dictionary $\bm D \in \R^{D\times K}$ and a decision function $ \hat{y} = \phi (\bm X_n, \bm D; \bm \theta)$ for fine-grained classification. Specifically, $\bm D=[\bm d_{1}, \bm d_{2},..., \bm d_{K}]$ and each column vector $\bm d_k$ represents an object part concept. And $\bm \theta$ are the parameters of $\phi(\cdot)$. $\phi(\cdot)$ thus takes both the feature maps $\bm X_n$ and the part dictionary $\bm D$ to predict the labels $y_n$. We now present an overview of our model, as illustrated in Fig.\ \ref{fig:structure}. Without loss of clarity, we sometimes drop the subscript $n$.

Specifically, we assume the function $\phi (\bm X, \bm D; \bm \theta)$ can be further decomposed into three parts. 
\begin{itemize}\vspace{-0.6em}
    \item \textbf{Part Segmentation}. A soft part assignment map $\bm Q \in \R^{K\times H \times W}$ is created by comparing the feature map $\bm X$ to the part dictionary $\bm D$. This is achieved by using a grouping function $g(\cdot)$ such that $\bm Q = g(\bm X, \bm D; \bm \theta_g)$. \vspace{-0.4em}
    \item \textbf{Region Feature Extraction and Attribution}. Based on the assignment map $\bm Q$ and part dictionary $\bm D$, region features $\bm Z \in \R^{D\times K}$ are pooled from the feature maps $\bm X$. We further compute an attention vector $\bm a \in \R^K$, where each element provides the importance score for a part segment. Formally, $[\bm Z, \bm a] = f(\bm X, \bm Q, \bm D; \bm \theta_f)$. \vspace{-0.4em}
    \item \textbf{Attention Based Classification}. The region features $\bm Z$ are re-weighted by region attention $\bm a$, followed by a linear classifier for the decision of $y$. This is realized by $h(\cdot)$ where $\hat{y} = h(\bm Z, \bm a; \bm \theta_c)$.
\end{itemize}

\noindent \textbf{Regularization of Part Occurrence}. Before we describe our design of $g(\cdot)$, $f(\cdot)$ and $h(\cdot)$, let us look at the major challenge of learning. Since the only supervisory signal is $y$, it is challenging to make sure that the dictionary $\bm D$ can capture meaningful object parts. Our key assumption is that we can regularize the learning by enforcing a prior distribution for the occurrence of each part $\bm d_k$ within a set of image features $\bm X_{1:N}$. Specifically, given $\bm X_{1:N}$, with slightly abusing of notation, we denote $p(\bm d_k | \bm X_{1:N})$ as the conditional probability of part $\bm d_k$ occurring in the set $\bm X_{1:N}$. We assume that $p(\bm d_k | \bm X_{1:N})$ follows a U-shaped distribution $\hat{p}(\bm d_k)$ that acts like a probabilistic binary switch, where we can control the probability of ``on'' and ``off''. For example, on the CUB-200 birds dataset, all bird parts are presented in most of the bird images, such that the switch is almost always on. In contrast, on the more challenging iNaturalist dataset, an object part is only activated for a certain number of categories, and thus the switch is likely to activate only for some of the images.

\subsection{Part Segmentation and Regularization}
We now present the details for part segmentation and how to regularize the occurrence of parts. 

\noindent \textbf{Part Assignment}. We adopt a similar projection unit proposed in previous work~\cite{li2018beyond,chen2019graph}. More precisely, let $q_{ij}^k$ be an element in the assignment matrix $\bm Q$, where $i, j$ index the 2D position and $k$ indexes the parts. $q_{ij}^k$ indicates the probability for a feature vector $\bm x_{ij}\in \R^D$ at position $(i, j)$ on $\bm X$ to be assigned to the $k$-th part $\bm d_k$ in $\bm D$. $q_{ij}^k$ is computed as
\begin{equation}
    \small
    q_{ij}^k = \frac{\exp(-\lVert (\bm x_{ij} - \bm d_k)/\sigma_k \rVert^2_2 / 2)}{\sum_k \exp(-\lVert (\bm x_{ij} - \bm d_k)/\sigma_k \rVert^2_2 / 2)}, \label{eq:4}
\end{equation}
where $\sigma_k \in (0, 1)$ is a learnable smoothing factor for each part $\bm d_k$. Due to the softmax normalization, we have $q_{ij}^k>0$ and $\sum_k q_{ij}^k=1$. Moreover, We assemble all assignment vectors $\bm q_{ij} \in \R^K$ into our part assignment map $\bm Q \in \R^{K\times H \times W}$.

\noindent \textbf{Part Occurrence}. Given the part assignment map, our next step is to detect the occurrence of each part $\bm d_k$. A simple part detector can be implemented using a max pooling operation over the $k$-th assignment map $\bm Q^k = [q_{ij}^k]$. However, we found it beneficial to ``smooth'' the assignment map before pooling, e.g., using a Gaussian kernel with a small bandwidth. This smoothing operation helps to eliminate outliers on the feature map. Our part detector is thus defined as $t_k = \max_{ij}\ \mathcal{G} * \bm Q^k$, where $\mathcal{G}$ is a 2D Gaussian kernel and $*$ is the convolution operation. $t_k$ lies in the range of $(0, 1)$. Furthermore, the outputs of $k$ part detectors are concatenated into an occurrence vector $\bm \tau = [t_1, t_2, ..., t_K]^T \in (0, 1)^K$ for all parts.

\noindent \textbf{Regularization of Part Occurrence}. Our key idea is to regularize the occurrence of each part. This is done by enforcing the empirical distribution of part occurrence to align with a U-shaped prior distribution. More precisely, given a set of $N$ samples, e.g., those used in a mini-batch sampled from the full dataset, we first estimate the empirical distribution $p(\bm d_k | \bm X_{1:N})$ by concatenating all occurrence vectors $\tau_n, n=1,2, ..., N$ into an matrix $\bm T = [\bm \tau_1, \bm \tau_2, ..., \bm \tau_N] \in (0, 1)^{K \times N}$. Moreover, we assume a prior distribution $\hat{p}(\bm d_k)$ is known, e.g., a Beta distribution. We propose to use 1D Wasserstein distance, also known as Earth-Mover distance to align $p(\bm d_k | \bm X_{1:N})$ and the prior $\hat{p}(\bm d_k)$, given by
\begin{equation*}
    \small
    W(p(\bm d_k | \bm X_{1:N}), \hat{p}(\bm d_k)) = \int_{0}^{1} |F^{-1}(z) - \hat{F}^{-1}(z)| d z, 
\end{equation*}
where $F(\cdot)$ and $\hat{F}(\cdot)$ are the Cumulative Distribution Functions (CDFs) for the empirical and prior distribution. And $z$ spans the interval of $[0, 1]$. 

During mini-batch training, the Wasserstein distance can be approximated by replacing the integration with a summation over the samples within the mini-batch, leading to the L$1$ distance between $F^{-1}$ and $\hat{F}^{-1}$. In practice, we find that it is helpful to rescale the inverse of the CDFs using a logarithm function, which improves the stability of training.
\begin{equation}
\small
\begin{split}
    \mathcal{L}_W = \frac{1}{N} \sum_{n=1}^N |
    \log ([\bm \tau_{k}^*]_i + \epsilon)
    - \log (\hat{F}^{-1}\left(\frac{2n-1}{2N}\right) + \epsilon)|, \label{eq:13}
\end{split}
\end{equation}
where $\bm \tau_k^*$ is the sorted version (in ascending order) of the $k$-th row vector of $\bm T$ (size N) and $[\bm \tau_k^*]_i$ is the $i$-th element of $\bm \tau_k^*$. $\epsilon$ is a small value added for numerical stability. Using the logarithm rescaling overcomes a gradient vanishing problem introduced by the softmax function in Eq.\ \ref{eq:4}. Even if a part $\bm d_k$ is far away from all feature vectors in the current mini-batch, i.e., with a small value of $q_{ij}^k$ in Eq.\ \ref{eq:4}, $\bm d_k$ can still receive non-zero gradients due to the rescaling.  

We note that there are different approaches to align two 1D distributions. We have previously experimented with Cram\`{e}r-von Mises criterion by shaping the CDFs as proposed in~\cite{bejnordi2019batch}. However, we found that our choice of 1D Wasserstein produces more robust results across datasets.

\subsection{Region Feature Extraction and Attribution}
Given the part assignment, our next step is to pool features from each region. This is done by using a nonlinear feature encoding scheme~\cite{li2018beyond, jegou2010aggregating, perronnin2010improving, arandjelovic2013all}, given by
\begin{equation}
    \small
    \bm z_k' = \frac{1}{\sum_{ij} q_{ij}^k} \sum_{ij} q_{ij}^k (\bm x_{ij} - \bm d_{k}) / \sigma_k,\ \  \bm z_k = \frac{\bm z_k'}{\lVert \bm z_k' \rVert_2}. \label{eq:5}
\end{equation}
$\bm z_k$ is thus the region feature from pixels assigned to part $\bm d_k$. By combining $\bm z_k$ together, we obtain our region feature set $\bm Z = [\bm z_1, \bm z_2, ..., \bm z_K] \in \R^{D \times K}$ from the input feature maps. We further transform $\bm Z$ using a sub-network $f_z$ that has several residual blocks, where each one is a bottleneck block consisting of three 1x1 convolutions with Batch Normalization~\cite{ioffe2015batch} and ReLU. The transformed features are thus $f_z(\bm Z)$. 

Moreover, an attention module is attached on top of $\bm Z$ to predict the importance of each region. This is realized by a sub-network $f_a$, given by $\bm a = \mbox{softmax}(f_a(\bm Z^T))$,
where $f_a$ consists of multiple 1x1 convolutions with Batch Normalization and ReLU in-between. The result attention $\bm a \in \R^K$ is further used for classification.

\subsection{Attention Based Classification}
Finally, we re-weight the transformed region features $f_z(\bm Z)$ using the attention vector $\bm a$, followed by a linear classifier. Therefore, the final prediction is given by
\begin{equation}
    \small
    \hat{y} = \mbox{softmax}( \bm W f_z(\bm Z) \bm a)
\end{equation}
where $\bm W\in \R^{C\times D}$ is the weights of a linear classifier for $C$-way classification. Note that the attention $\bm a$ serves as a modulator of the region features $\bm Z$. Thus, large values in $\bm a$ suggest a more important region for classification. 

\noindent \textbf{Pixel Attribution}. Given the attention $\bm a$, we can easily back-track the contribution of each pixel on the feature map. This can be done by using $\mathcal{Q}^T \bm a$, where $\mathcal{Q} \in \R^{K \times HW}$ is simply a reshaped version of the part assignment map $\bm Q$.

\subsection{Implementation}
We present our implementation details on loss functions, network architecture, as well as training and inference. 

\noindent \textbf{Loss Function}. Our model was trained by minimizing cross entropy loss for classification plus the 1D Wasserstein distance based regularization loss in Eq.\ \ref{eq:13} for part regularization. We varied the weights balancing the loss terms and the prior Beta distribution used for Wasserstein distance during our experiments.

\noindent \textbf{Network Architecture}. We replaced the last convolutional block of a baseline CNN (ResNet101~\cite{he2016deep} ) with our proposed module. We roughly matched the number of parameters in our final model to the baseline.

\noindent \textbf{Training and Inference}. We used standard mini-batch SGD for all datasets. The hyper-parameters were chosen differently across datasets due to varied tasks and will be discussed in the experiments. We applied data augmentation including random crop, random horizontal flip and color jittering, and adopted learning rate decay as in~\cite{he2016deep}. The convolutional layers in our models were initialized from ImageNet pre-trained models. The new parameters, including the part dictionary, were randomly initialized following~\cite{he2015delving}. All parameters were jointly learned on the target dataset. For all experiments, we report results using a single center crop unless further notice.

%% file: src/exp.tex
\section{Experiments and Results}
We now describe our experiments and discuss the results. We first introduce datasets and metrics used in our experiments. We then present our experiments and results on individual dataset, followed by an ablation study. For all experiments, results are reported on both accuracy and interpretability and compared against latest methods. 

\noindent \textbf{Datasets}. Three fine-grained recognition datasets are considered in our experiments, including CelebA~\cite{liu2015deep}, CUB-200-2011~\cite{wah2011caltech} and iNaturalist 2017~\cite{van2018inaturalist}. These datasets span over a range of tasks and sizes. CelebA is a medium scale dataset for facial attribute recognition and facial landmark detection. CUB-200 is a small scale dataset for bird species recognition that also comes with bird keypoint annotations. Finally, iNaturalist 2017 is a large-scale dataset for fine-grained species recognition and detection, with over 5000 categories spanning from mammals to plants.

\noindent \textbf{Evaluation Metric}. We evaluate both accuracy and interpretability of our model for fine-grained visual recognition. For accuracy, we report the standard instance level or average class accuracy as previously considered for fine-grained classification. As a proxy of interpretability, we measure object part localization error using annotated object landmarks, since our model is designed to discover object parts. This localization error has been previously considered for part segmentation models, such as Hung et al.\ \cite{hung2019scops}. For the dataset, e.g., iNaturalist 2017 that does not come with part annotations, we follow the protocol of Pointing Game~\cite{zhang2018top} and report object localization error using the annotated object bounding boxes. Pointing Game has been widely used to evaluate interpretable deep models~\cite{zhang2018top,selvaraju2017grad,petsiuk2018rise}.   

Concretely, part localization errors are reported on CelebA and CUB-200. Following a similar protocol in~\cite{hung2019scops}, we convert our assignment maps to a set of landmark locations by learning a linear regression model. The regression model maps the 2D geometric centers of part assignment maps into 2D object landmarks. The predicted landmarks are compared against ground-truth on the test set. We report normalized mean L2 distance between the prediction and ground-truth. For iNaturalist 2017, Pointing Game results are reported. We calculate the error rate by counting the cases where the peak location of our output attention map lies outside the ground-truth object bounding boxes. 

\subsection{Results on CelebA}
\noindent \textbf{Dataset}. CelebA~\cite{liu2015deep} is a facial attribute and landmark detection dataset that contains 202,599 celebrity face images collected from Internet. Each face image is annotated with 40 facial attributes and 5 landmark locations (eyes, noise and mouth corners). We consider two different splits of the data from~\cite{liu2015deep,hung2019scops}. The first split from~\cite{liu2015deep} includes 162,770, 19,867, and 19,962 images for training, validating and testing respectively, and is used to evaluate facial attribute recognition. Faces in this split are aligned to image center. The second split from~\cite{hung2019scops} has 45,609 images for training, 5,379 images for fitting the linear regressor and 283 images for testing. This split is used to report part localization error. Faces are not aligned in this split. 
 
\begin{table}[t]
\centering \small
\begin{tabular}{l|c}
Method   & Acc (\%) $\uparrow$ \\ \hline
LNet+ANet~\cite{liu2015deep} & 87.0     \\ \hline
MOON~\cite{rudd2016moon} & 90.9   \\ \hline
Lu et al.\ \cite{lu2017fully} & 91.0   \\ \hline
Hand et al.\ \cite{hand2017attributes} & 91.2   \\ \hline
Kalayeh et al.\ \cite{kalayeh2017improving} & 91.8   \\ \hline
He et al.\ \cite{he2018harnessing} & 91.8   \\ \hline
PS-MCNN~\cite{cao2018partially}  & \textbf{93.0}     \\ \hline \hline
ResNet101                        & 91.5     \\ \hline
\textbf{Ours}     & 91.5     \\ 
\end{tabular}
\vspace{0.1em}
\caption{Results of facial attribute recognition on CelebA dataset. Average class accuracy is reported.}\vspace{0.5em}
\label{table:celeba_1}
\begin{tabular}{c|c|c|c}
 &DFF & SCOPS & \textbf{Ours}  \\ \hline
Error (\%) $\downarrow$   & 31.3  & 15.0 & \textbf{8.4}    \\ 
\end{tabular}
\vspace{0.1em}
\caption{Results of facial landmark localization on CelebA dataset. Normalized L2 distance (\%) is reported.}
\label{table:celeba_2}\vspace{-1.5em}
\end{table}

\noindent \textbf{Implementation Details}. We trained two models on the two splits using the same architecture. We attached a separate attention-based binary classification head for each facial attribute, as these attributes are not mutually exclusive. For attribute recognition, our model was trained on the training set and evaluated on test set. The validation set was used to select hyper-parameters. For landmark localization, we followed the training procedure from~\cite{hung2019scops}. Our models were trained using a learning rate of 5e-3 with batch size of 32 and a weight decay of 1e-4 for 30 epochs. We set the weights between the two loss terms to 10:1 and used a prior Beta distribution with $\alpha$=1 and $\beta$=1e-3 (close to Bernoulli with $p=1$). All input images were resized to $256\times 256$ and fed into the model without cropping. A dictionary of 9 parts was used. When reporting the part localization error, we normalize the L2 distance by the inter-ocular distance~\cite{hung2019scops}.

\begin{figure}[t]
\centering
\bgroup
\def\arraystretch{0.2}
\begin{tabular}{@{}c@{}c@{}c@{}c@{}}
\includegraphics[width=1.8cm]{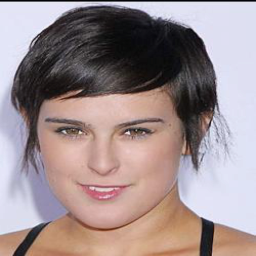}&
\includegraphics[width=1.8cm]{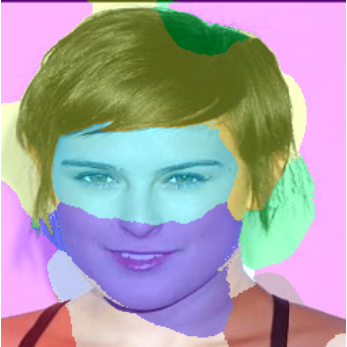}&
\includegraphics[width=1.8cm]{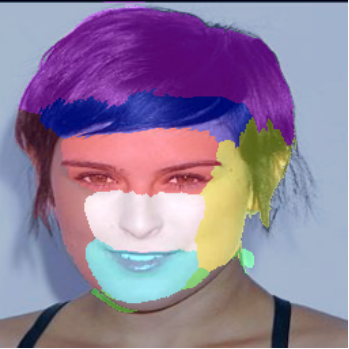}&
\includegraphics[width=1.8cm]{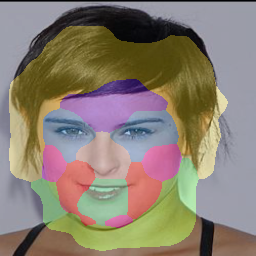}\\
\includegraphics[width=1.8cm]{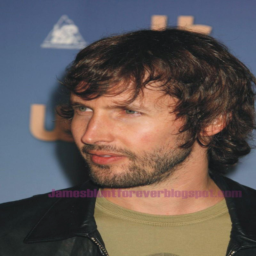}&
\includegraphics[width=1.8cm]{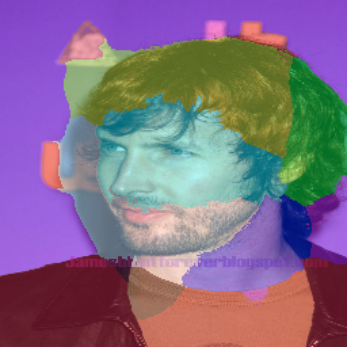}&
\includegraphics[width=1.8cm]{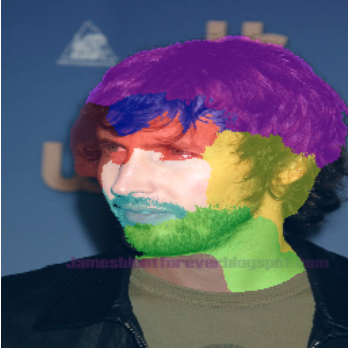}&
\includegraphics[width=1.8cm]{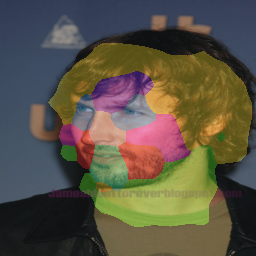}\\
\includegraphics[width=1.8cm]{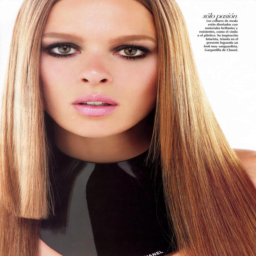}&
\includegraphics[width=1.8cm]{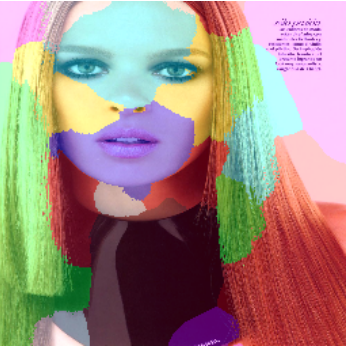}&
\includegraphics[width=1.8cm]{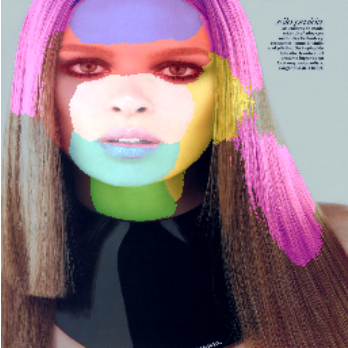}&
\includegraphics[width=1.8cm]{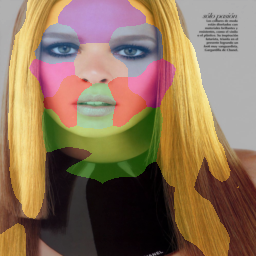}\\
\includegraphics[width=1.8cm]{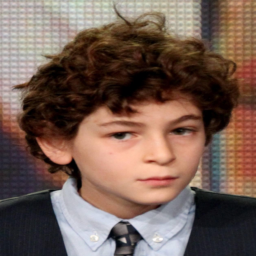}&
\includegraphics[width=1.8cm]{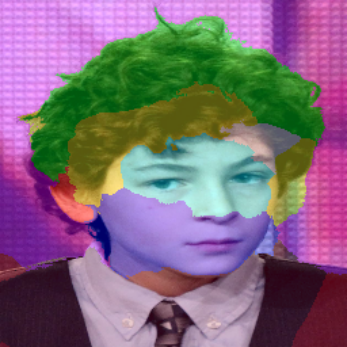}&
\includegraphics[width=1.8cm]{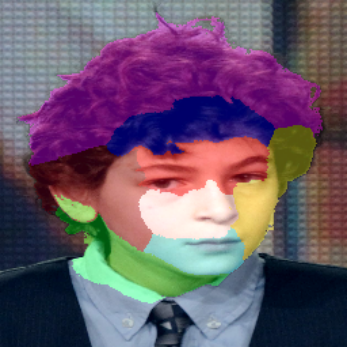}&
\includegraphics[width=1.8cm]{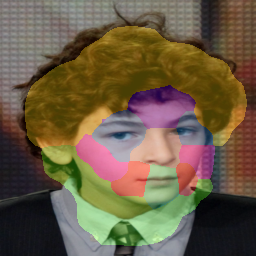}\\
\\
\ \ &
\ \ &
\ \ &
\ \ \\
\small Input &\small DFF &\small SCOPS &\small Ours\\
\end{tabular}
\egroup
\vspace{0.5em}
\caption{Visualization of part assignment maps on CelebA. From left to right: input images, DFF results~\cite{collins2018}, SCOPS results~\cite{hung2019scops} and our results. Our results are better aligned with the facial parts.}
\label{fig:celeb_comp} \vspace{0.5em}
\centering
\bgroup
\def\arraystretch{0.2}
\begin{tabular}{@{}c@{}c@{}c@{}c@{}}
\includegraphics[width=1.8cm]{figs/celeba/inputs/1.png}&
\includegraphics[width=1.8cm]{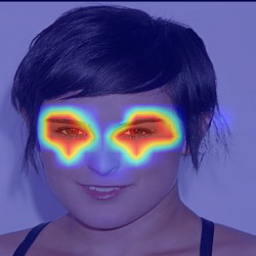}&
\includegraphics[width=1.8cm]{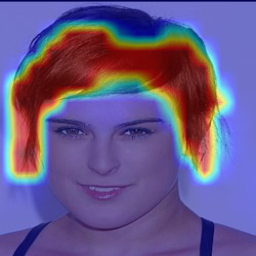}&
\includegraphics[width=1.8cm]{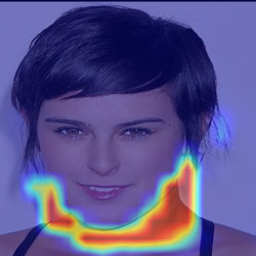}\\
\includegraphics[width=1.8cm]{figs/celeba/inputs/2.png}&
\includegraphics[width=1.8cm]{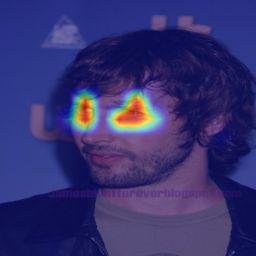}&
\includegraphics[width=1.8cm]{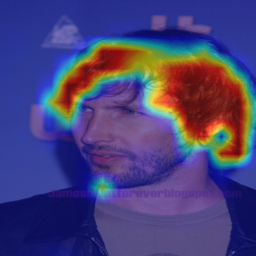}&
\includegraphics[width=1.8cm]{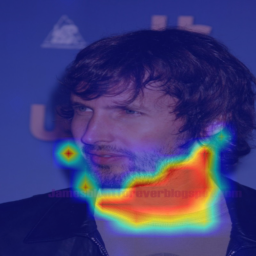}\\
\includegraphics[width=1.8cm]{figs/celeba/inputs/3.png}&
\includegraphics[width=1.8cm]{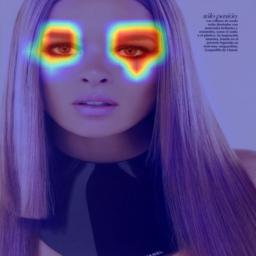}&
\includegraphics[width=1.8cm]{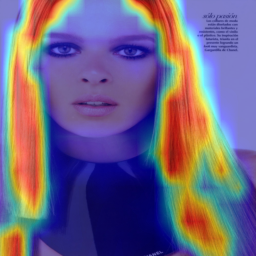}&
\includegraphics[width=1.8cm]{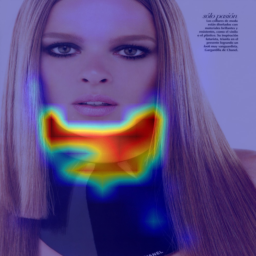}\\
\includegraphics[width=1.8cm]{figs/celeba/inputs/4.png}&
\includegraphics[width=1.8cm]{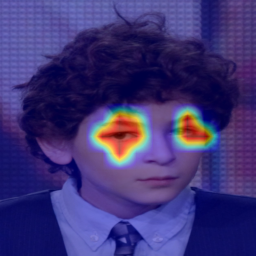}&
\includegraphics[width=1.8cm]{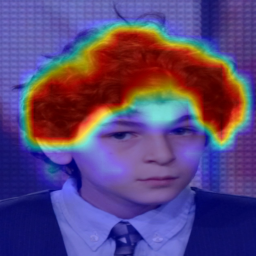}&
\includegraphics[width=1.8cm]{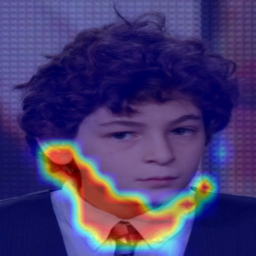}\\
\\
\ \ &
\ \ &
\ \ &
\ \ \\
\small Input&\small Narrow &\small Black &\small Wearing\\
\small \ &\small Eyes&\small Hair&\small Necktie\\
\end{tabular}
\egroup
\vspace{0.5em}
\caption{Visualization of attention maps from our model on CelebA. From left to right: input images, attention maps corresponding to different attributes. Our model is able to identify regions that are discriminative for facial attributes.}
\label{fig:celeb_att}\vspace{-1.5em}
\end{figure}

\noindent \textbf{Recognition Results (Accuracy)}. Our results on attribute recognition are summarized in Table.\ \ref{table:celeba_1}. We compare our results with state-of-the-art methods~\cite{liu2015deep, rudd2016moon, lu2017fully, hand2017attributes, kalayeh2017improving, he2018harnessing, cao2018partially} as well as a baseline ResNet101 (pre-trained on ImageNet). Surprisingly, the baseline ResNet101 already achieves similar or even higher accuracy in comparison to many previous methods, including those require auxiliary face parsing~\cite{kalayeh2017improving, he2018harnessing}. Our model performs on par with the strong ResNet101 baseline. The only method that is significantly better than our model and ResNet101 baseline is~\cite{cao2018partially}, which uses extra labels of face identities. To summarize, our model achieves state-of-the-art accuracy.

\noindent \textbf{Localization Results (Interpretability)}. We further evaluate facial landmark localization results, as shown in Table~\ref{table:celeba_2}. Our results are compared to the latest methods of DFF~\cite{collins2018} and SCOPS~\cite{hung2019scops}. DFF performs non-negative matrix factorization on the feature maps of a pre-trained CNN (VGG\footnote{Using ResNet leads to worse results of DFF as reported in~\cite{collins2018}.}) to generate the part segmentation. SCOPS explores spatial coherence, rotation invariance, semantic consistency and visual saliency in self-supervised training for object part segmentation. Our model outperforms both methods by a significant margin in localization error, achieving a 6.6\% and 21.9\% error reduction when compared to SCOPS and DFF, respectively. These results suggest that our model can localize facial landmarks with high accuracy, thus supporting the interpretability of our model.

\noindent \textbf{Visualization}. Our model achieves state-of-the-art results on facial attribute recognition and provides new capacity for facial landmark localization. We further visualize the assignment maps from our model and compare them to those from DFF~\cite{collins2018} and SCOPS~\cite{hung2019scops} in Fig.\ \ref{fig:celeb_comp}. Moreover, we display the attention maps from our model in Fig.\ \ref{fig:celeb_att}. Note that our attention maps are attribute specific, as we used a separate classification head for each attribute. These qualitative results show that our model is able to segment faces into meaningful part regions (e.g., hair, forehead, eyes, nose, mouth and neck), and to attend to those regions that are discriminative for attribute recognition (e.g., eye regions for ``narrow eyes'' or hair regions for ``black hair'').

\begin{table}[t]
\centering  \small
\begin{tabular}{l|c}
Method         & Acc (\%) $\uparrow$ \\ \hline
STN~\cite{jaderberg2015spatial}            & 84.1          \\ \hline
Kernel Pooling~\cite{cui2017kernel} & 86.2          \\ \hline
MA-CNN~\cite{zheng2017learning}         & 86.5          \\ \hline
PC-DenseNet161~\cite{dubey2018pairwise}    & 86.9          \\ \hline
KERL~\cite{chen2018knowledge}           & 87.0          \\ \hline
DFL-CNN~\cite{wang2018learning}        & \textbf{87.4}          \\ \hline
NTS-Net~\cite{yang2018learning}        & \textbf{87.5}          \\ \hline
DCL~\cite{chen2019destruction}            & \textbf{87.8}          \\ \hline
TASN~\cite{zheng2019looking}           &\textbf{87.9}          \\ \hline \hline
ResNet101               & \textbf{87.7}          \\ \hline
\textbf{Ours}           & \textbf{87.3}          \\ 
\end{tabular}
\vspace{0.1em}
\caption{Results of bird species recognition on CUB-200-2011. Instance level accuracy is reported.}\vspace{0.5em}
\label{table:cub_1}
\centering  \small
\begin{tabular}{l|c|c|c}
Method & CUB-001 & CUB-002 & CUB-003  \\ \hline
DFF    & 22.4   & 21.6   & 22.0    \\ \hline
SCOPS  & 18.5    & 18.8   & 21.1    \\ \hline
\textbf{Ours}   & \textbf{15.6}   & \textbf{15.9}   & \textbf{13.8}    \\ 
\end{tabular}
\vspace{0.1em}
\caption{Results of landmark localization errors on CUB-200-2011. Normalized L2 distance (\%) is reported.}
\label{table:cub_2}\vspace{-1.5em}
\end{table}

\begin{figure}[t]
\centering
\bgroup
\def\arraystretch{0.0}
\begin{tabular}{@{}c@{}c@{}c@{}c@{}}
\includegraphics[width=1.8cm]{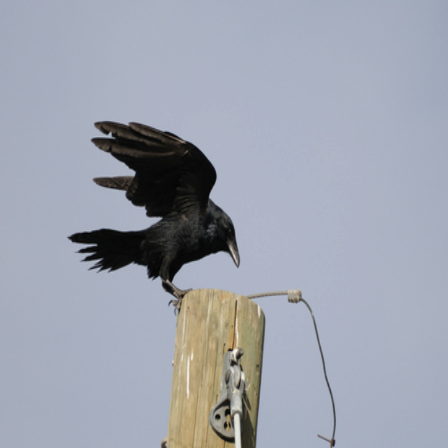}&
\includegraphics[width=1.8cm]{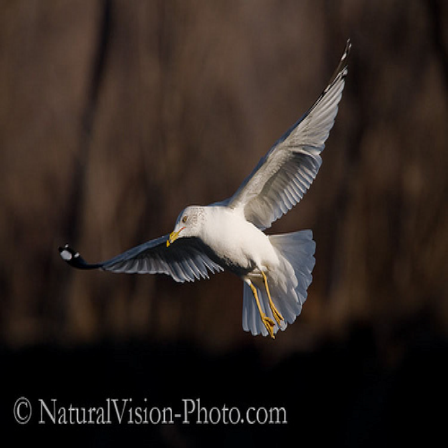}&
\includegraphics[width=1.8cm]{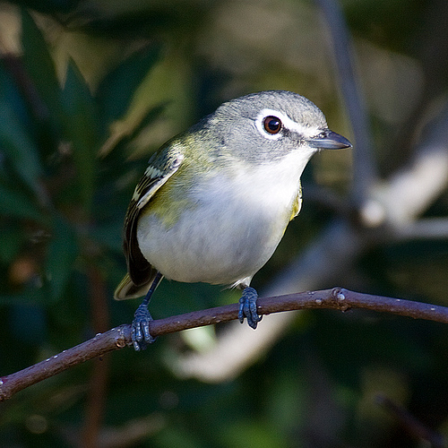}&
\includegraphics[width=1.8cm]{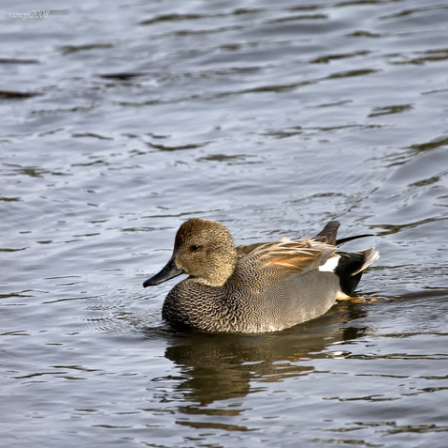}\\
\includegraphics[width=1.8cm]{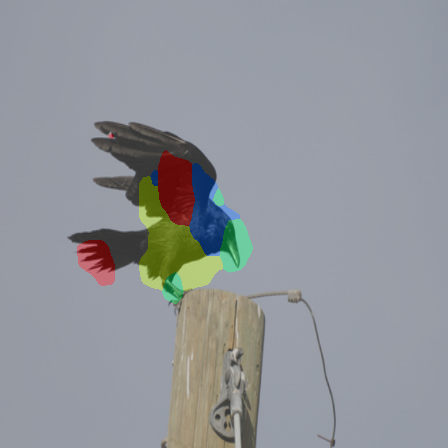}&
\includegraphics[width=1.8cm]{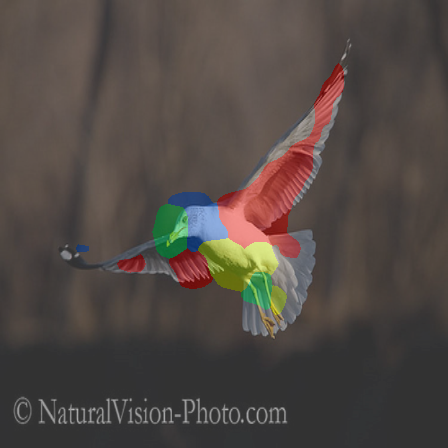}&
\includegraphics[width=1.8cm]{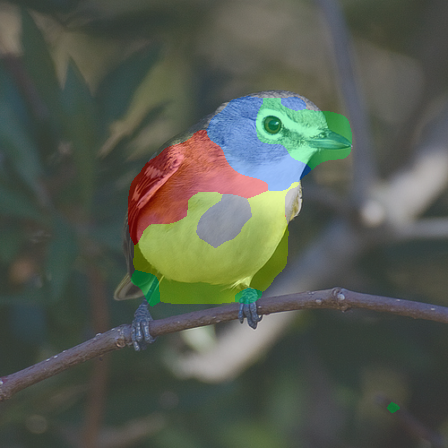}&
\includegraphics[width=1.8cm]{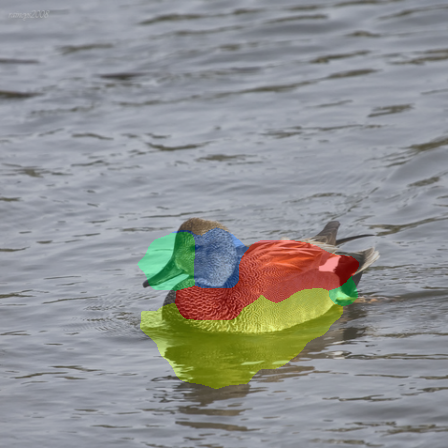}\\
\includegraphics[width=1.8cm]{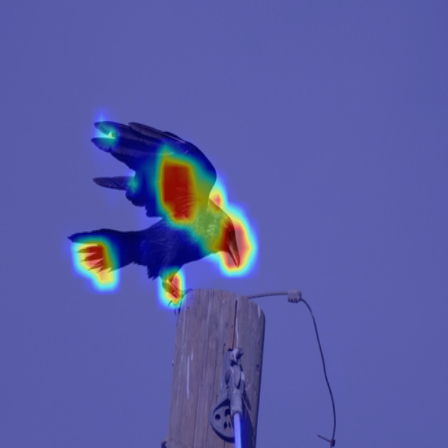}&
\includegraphics[width=1.8cm]{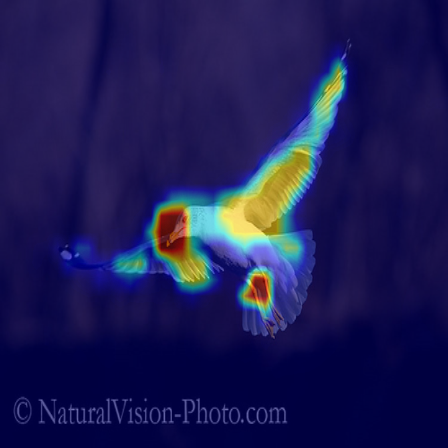}&
\includegraphics[width=1.8cm]{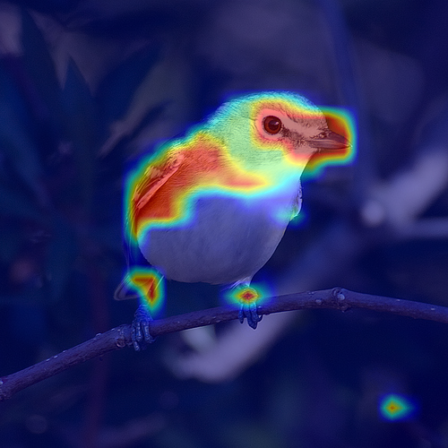}&
\includegraphics[width=1.8cm]{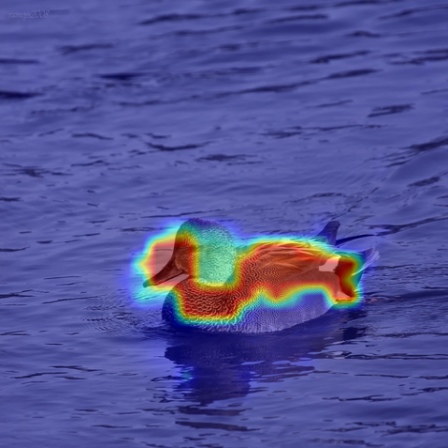}\\
\ \ &
\ \ &
\ \ &
\ \ \\
\vspace{0.5em}
\end{tabular}
\egroup
\caption{Visualization of sample assignment and attention maps from our model on CUB-200-2011 test set. From top to bottom: input images, assignment maps and attention maps. Our method can consistently identify body parts under different poses.}\vspace{-1em}
\label{fig:cub_vis}
\end{figure}

\subsection{Results on CUB-200-2011}
\noindent
\textbf{Dataset}. Caltech-UCSD Birds-200-2011~\cite{wah2011caltech} (CUB-200) is a small scale dataset for fine-grained bird species recognition. CUB-200 contains 5,994/5,794 images for training/test from 200 different bird species. Each image is annotated with a species label, 15 bird landmarks and a bounding box of the bird. 

\noindent
\textbf{Implementation Details}. We trained a single model for both classification and landmark localization using a learning rate of 1e-3, batch size of 32 and a weight decay of 5e-4 for 150 epochs. We set the weights between the two loss terms to 2:1, used a prior distribution the same as CelebA and a dictionary of 5 parts. We resized the input images by scaling the shortest side to 448 and randomly crop a region of 448x448 for training. When reporting the part localization error, we normalized the L2 distance using the size of the bird's bounding box, similar to~\cite{hung2019scops}. 

\noindent \textbf{Recognition Results (Accuracy)}. We present our results of recognition accuracy and compare them to state-of-the-art methods~\cite{jaderberg2015spatial,cui2017kernel,zheng2017learning,dubey2018pairwise,chen2018knowledge,wang2018learning,yang2018learning, chen2019destruction,zheng2019looking} in Table.~\ref{table:cub_1}. Again, the baseline ResNet101 already achieves state-of-the-art results on CUB-200. Our model is slightly worse than ResNet101 (-0.4\%) and performs on par with previous part-based models like MA-CNN~\cite{zheng2017learning}.

\noindent \textbf{Localization Results (Interpretability)}. Moreover, we evaluate part localization error and compare our results to DFF~\cite{collins2018} and SCOPS~\cite{hung2019scops}. To make a fair comparison, we report the errors on the first three categories following~\cite{hung2019scops}, as shown in Table~\ref{table:cub_2}. Again, our model significant reduces the localization error (2.9\%--6.2\%). When fitting with all 200 categories, our model achieves an average localization error of $11.51$\%. These results provide further evidences towards the interpretability of our model. 

\begin{figure*}[t]
\centering
\includegraphics[width=0.9\linewidth]{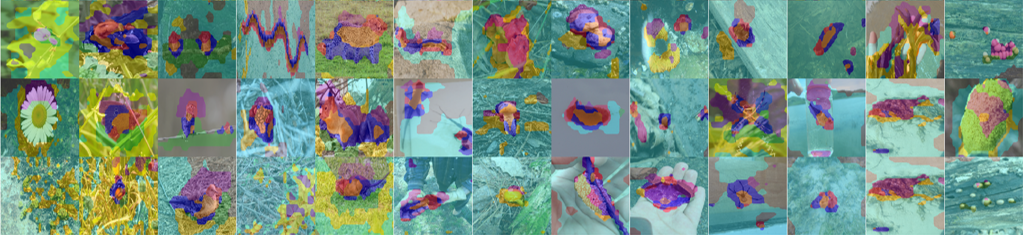}
\caption{Visualization of sample assignment maps on iNaturalist 2017 test set. Each column comes from one super category.}\vspace{-0.5em}
\label{fig:inat_ass}
\end{figure*}

\begin{figure*}[h]
\centering
\includegraphics[width=0.9\linewidth]{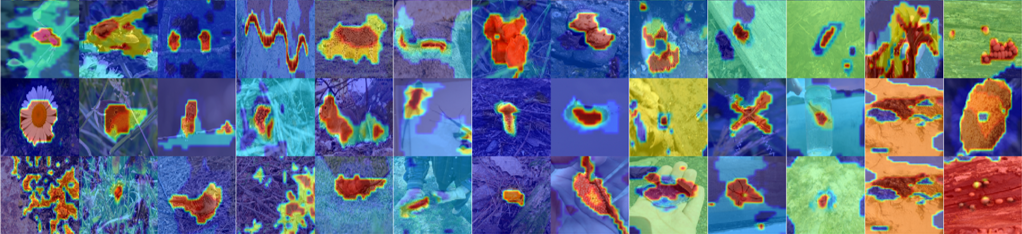}
\caption{Visualization of sample attention maps on iNaturalist 2017 test set. Each column comes from one super category.}\vspace{-0.5em}
\label{fig:inat_att}
\end{figure*}

\noindent \textbf{Visualization}. We also visualize the assignment maps and the attention maps from our model, as presented in Fig.\ \ref{fig:cub_vis}. Our model demonstrates the ability to discover coherent parts of the birds (e.g, beak/legs, head, wings/tail, body) and select the important regions (beak/legs and wings/tails) to recognize the species.

\subsection{Results on iNaturalist 2017}
\noindent \textbf{Dataset}.
iNaturalist 2017~\cite{wah2011caltech} is a large-scale dataset for fine-grained species recognition. It contains 579,184 and 95,986 for training and testing from 5,089 species organized into 13 super categories. Some images also come with bounding box annotations of the object. Since the dataset does not include part annotations, we report Pointing Game results to evaluate interpretability of our model. This dataset is very challenging for mining meaningful object parts, as the objects in different super-categories have drastically different visual appearance (e.g., plants vs.\ mammals). 

\noindent
\textbf{Implementation Details}.
We trained a single model for classification and localization. Our model was trained using a learning rate of 1e-3 with batch size of 128 and a weight decay of 5e-4 for 75 epochs. During training, we resize the input images by scaling the shortest side to 320 and randomly crop a region of 224x224. We set the weights between the two loss terms to 10:1. A dictionary of 8 parts was used and a prior Beta distribution with $\alpha$=2e-3 and $\beta$=1e-3 was considered. We also explored fully convolutional test by feeding the full image (shortest side of 320) into the model. 

\begin{table}[ht]
\centering \small
\begin{tabular}{l|c}
Method              & Acc (\%) $\uparrow$ \\ \hline
SSN                 & 65.2     \\ \hline
TASN                & \textbf{68.2}     \\ \hline \hline
ResNet101~\cite{zheng2019looking}           & 59.6    \\ \hline
ResNet101 (ours)          & 61.1    \\ \hline
\textbf{Ours}             & 64.8    \\ \hline
\textbf{Ours + FC test}           & 66.8    \\ 
\end{tabular}\vspace{0.1em}
\caption{Results of species classification on iNaturalist 2017. Instance level accuracy is reported. FC test: fully convolutional test.}\vspace{0.5em}\label{table:inat_acc}
\centering \small
\begin{tabular}{l|c}
Method         & Error (\%) $\downarrow$ \\ \hline
CAM~\cite{zhou2016learning} / Grad-CAM~\cite{selvaraju2017grad}   & 11.8       \\ \hline
Guided Grad-CAM~\cite{selvaraju2017grad}  & 8.2        \\ \hline
\textbf{Ours}    & \textbf{7.6}        \\ 
\end{tabular}\vspace{0.1em}
\caption{Results of Pointing Game on iNaturalist 2017. CAM/Grad-CAM \& Guided Grad-CAM use a ResNet101 model.} \vspace{-1.5em}\label{table:inat_pointing}
\end{table}

\noindent \textbf{Recognition Results (Accuracy)}. Table~\ref{table:inat_acc} summarizes our results and compares them to baseline ResNet101 models and the latest methods including SSN~\cite{recasens2018learning} and TASN~\cite{zheng2019looking}. Both SSN and TASN make use of attention-based upsampling to zoom in discriminative regions for classification. Unlike CelebA and CUB-200, the baseline ResNet101 result is much worse than state-of-the-art models (SSN and TASN). Our model improves the baseline ResNet101 by at least 3.7\%. Using test time augmentation (full convolutional testing) further boosts our results by 2\%. However, our model is still worse than TASN (-1.4\%). We speculate that a similar upsampling mechanism as SSN and TASN can be used by our model to further improve the accuracy. 

\noindent \textbf{Localization Results (Interpretability)}. Furthermore, we report Pointing Game results in Table~\ref{table:inat_pointing}. Our results are further compared to widely used saliency methods using the baseline ResNet101 model, including CAM/Grad-CAM~\cite{zhou2016learning,selvaraju2017grad} and Guided Grad-CAM~\cite{selvaraju2017grad}. Note that CAM and Grad-CAM are the same when visualizing features from the last convolutional layer of a ResNet. Our model achieves the lowest localization error (4.2\% and 0.6\% improvements for CAM/Grad-CAM and Guided Grad-CAM). Finally, the visualization of the assignment and attention maps are shown in Fig.\ \ref{fig:inat_ass} and Fig.\ \ref{fig:inat_att}.

\subsection{Ablation Study, Limitation and Discussion}

\begin{table}[t]
\centering \small
\begin{tabular}{l|c|c}
Method             & Acc (\%) $\uparrow$ & Error (\%) $\downarrow$ \\ \hline
w/o attention      & \textbf{91.5}        & \textbf{7.6}      \\ \hline
w/o regularization & \textbf{91.3}         & 12.3      \\ \hline
full model         & \textbf{91.5}         & 8.4      \\ 
\end{tabular}
\vspace{0.1em}
\caption{Ablation study on CelebA using the split from~\cite{hung2019scops}. Both recognition accuracy and localization error are reported.}\vspace{-1.5em}
\label{table:ablation}
\end{table}

\noindent \textbf{Ablation}. We conduct an ablation study on CelebA to evaluate our model components. Our study considers two variants, one without regularization and one without attention. Table~\ref{table:ablation} presents both recognition accuracy and localization error on the split from~\cite{hung2019scops}. The accuracy of all our variants are quite similar, yet our regularization largely improves the localization accuracy (3.9\%). Our model without attention has slightly better part localization performance, yet lacks the critical ability of region and pixel attribution compared to our full model. Our full model has small localization error for all landmarks--7.4\%, 7.5\%, 9.1\%, 9.3\% and 8.6\% for left eye, right eye, nose, left mouth corner and right mouth corner, respectively.

\noindent \textbf{Limitation and Discussion}. Many failure cases of our model were found on iNaturalist dataset, as shown in Fig.\ \ref{fig:inat_ass} and \ref{fig:inat_att}. Our model may fail to group pixels into part regions and sometimes produce incorrect saliency maps. We speculate these failure cases are produced by our prior Beta distribution. With over 5K fine-grained categories on iNaturalist, a single U-shaped distribution for all parts might fail to describe the occurrence of parts. Moreover, our model does not model the interaction between parts and requires a moderate to large batch size to estimate the empirical distribution of part occurrence. A promising future direction is thus to explore better priors of object parts.

%% file: src/conclusion.tex
\section{Conclusion}
We presented an interpretable deep model for fine-grained classification. Our model leveraged a novel prior of object part occurrence and integrated region-based part discovery and attribution into a deep network. Trained with only image-level labels, our model can predict an assignment map of object parts, an attention map of the part regions and the object label, demonstrating strong results for object classification and object part localization. We believe our model provides a solid step towards interpretable deep learning and fine-grained visual recognition.\\

\noindent \textbf{Acknowledgment}:
The authors acknowledge the support provided by the UW-Madison Office of the Vice Chancellor for Research and Graduate Education with funding from the Wisconsin Alumni Research Foundation. The authors also thank Fangzhou Mu for helpful suggestions on the writing.